\documentclass[runningheads]{llncs}%
\usepackage{eccv}
\usepackage{eccvabbrv}
\usepackage{graphicx}
\usepackage{booktabs}
\usepackage{stfloats} 
\usepackage{caption} 
\usepackage{changepage}
\usepackage{wrapfig}
\usepackage{hyperref}

\usepackage{orcidlink}
\usepackage[table]{xcolor}
\usepackage{threeparttable}
\usepackage{multirow}
\usepackage{tabularx}
\usepackage{float} 
\begin{document}

\title{From Gaze to Meaning: A Training-Free \\
AI Agent for Unified Grounding and Explanation} 

\titlerunning{Gaze Target Agent}

\author{Shayan Nasiriboukani\inst{1}\orcidlink{0009-0008-2035-5374} \and
Sara Atito\inst{1,2}\textsuperscript{$\star$}\orcidlink{0000-0002-7576-5791} \and
Mohammad Nezamipour\inst{1}\orcidlink{0009-0000-0632-3087} \and
Muhammad Awais\inst{1,2}\textsuperscript{$\star$}\orcidlink{0000-0002-1122-0709}}

\begingroup
\renewcommand{\thefootnote}{\fnsymbol{footnote}}
\footnotetext[1]{Corresponding authors.}
\endgroup

\authorrunning{S.~Nasiriboukani et al.}

\institute{Centre for Vision, Speech and Signal Processing (CVSSP), University of Surrey, UK \and
Surrey Institute for People-Centred AI, University of Surrey, UK
\email{\{s.nasiriboukani,sara.atito,muhammad.awais\}@surrey.ac.uk}\\
\email{monezamipoor@gmail.com} \\ \url{https://shayan137.github.io/gta-project-page/}}
\maketitle

\begin{abstract}
Understanding human attention is fundamental for scene interpretation, yet existing approaches often rely on heavily trained models that lack interpretability. Prior methods struggle to jointly reason about gaze targets, attended objects, and visual grounding without extensive supervision. 
To the best of our knowledge, this work introduces the first training-free \textbf{G}aze \textbf{T}arget \textbf{A}gent (GTA) for gaze-guided reasoning across tasks such as gaze target prediction, attention localization, and object identification.
This is achieved by leveraging pretrained vision–language models, augmenting them with visually guided prompts, and employing a memory-based retrieval strategy for high-uncertainty samples to improve performance without additional training.
We evaluate our approach using both quantitative metrics and qualitative results. Quantitatively, our method achieves state of the art performance on the GazeFollow and GazeHOI benchmarks. Qualitatively, our agent provides detailed semantic predictions, predicts the correct targets even when ground truth labels are wrong, and remains flexible without vocabulary constraints.
\end{abstract}

\section{Introduction}
\label{sec:intro}

Understanding where people direct their attention is central to visual scene interpretation \cite{NIPS2015_ec895663, Recasens_2017_ICCV, Chong_2018_ECCV}. Humans naturally organize their perception around gaze cues\cite{Birmingham2008SocialAttention}: we first identify who is present in the scene, then infer where each person is looking, and finally determine the object or region that is being attended to. This sequence of reasoning allows observers to interpret the focus of attention within a scene and to identify elements that are likely to be important or behaviorally relevant.

Beyond simply locating visual targets, gaze cues provide rich social information that supports higher-level reasoning about human behavior. By analyzing where individuals direct their gaze, observers can infer intentions, anticipate
\begin{wrapfigure}[14]{r}{0.5\textwidth}
    \centering
    \includegraphics[width=0.5\textwidth]{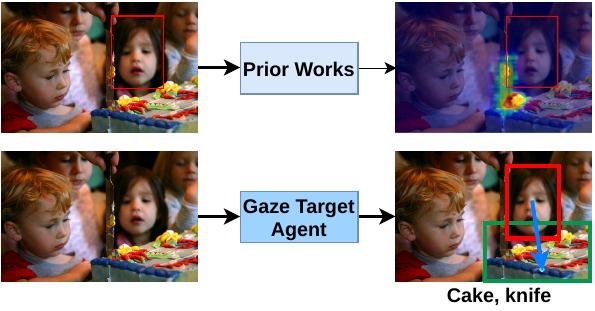}
    \small
    \caption{
    Comparison with prior approaches: prior methods predict gaze heatmaps indicating where a person is looking, whereas our approach not only estimates gaze but also grounds the attended object and identifies it.}
\label{fig:intro}
\end{wrapfigure}
 actions, and understand interactions between people and objects. Furthermore, shared or coordinated gaze among multiple individuals can reveal joint attention, which is a key signal of communication, collaboration, and social engagement\cite{Chong_2020_CVPR, Tafasca_2023_ICCV}. As a result, modeling gaze and attention has become an important component in computer vision systems that aim to interpret complex human-centered scenes. 
 
Despite the rapid progress of vision language models \cite{Qwenvl3,zhu2025internvl3exploringadvancedtraining,bai2025qwen25vltechnicalreport,molmo}, most systems still reason about scenes without explicitly modeling human attention. Existing approaches typically treat gaze estimation, object grounding, and explanation as separate problems, often relying on task-specific supervision and training pipelines that do not generalize well across domains. As a result, models may produce semantically fluent descriptions that are only weakly grounded in the visual evidence, and object hallucination remains a persistent challenge \cite{chen2024multiobjecthallucinationvisionlanguagemodels, leng2023mitigatingobjecthallucinationslarge}.

In this work, we investigate whether an AI agent can leverage gaze as a guiding signal for visual reasoning within a fully training-free framework, and formulate this setting as a gaze target agent. We build a simple gaze-in-the-loop pipeline(Fig.~\ref{fig:intro}) in which a head detector\cite{yaseen2024yolov8indepthexplorationinternal} identifies people and a zero-shot gaze estimator\cite{Ryan_2025_CVPR} predicts where they are looking. The predicted gaze is rendered as an arrow and used as a visual prompt for a vision language model\cite{Qwenvl3, an2025llava}, which selects an object label from a predefined vocabulary\cite{NEURIPS2024_dbeb7e62} and estimates prediction uncertainty, enabling reasoning about attended objects and shared attention. When uncertainty is high, the agent retrieves visually similar examples from memory and conditions the model on them to refine its prediction, improving robustness without updating any parameters.
Importantly, gaze also serves as a grounding signal by aligning predicted attention with candidate object regions, thereby reducing object hallucination and producing more robust and interpretable predictions.

We evaluate our agent on GazeFollow~\cite{NIPS2015_ec895663} and GazeHOI~\cite{NEURIPS2024_dbeb7e62}, achieving state of the art performance on gaze target prediction across both datasets. In addition, we compare several grounding models on GazeHOI, which provides object-level bounding boxes, to identify the most reliable grounding component for gaze-guided reasoning.
Beyond quantitative results, we conduct qualitative analyses to examine the semantic capabilities of our gaze target agent. Our model produces richer semantic predictions of attended objects, correctly identifies gaze targets in cases where ground truth annotations are inaccurate, and demonstrates flexible predictions when vocabulary constraints are removed. These results highlight the ability of a training-free gaze target agent to integrate gaze estimation, grounding, uncertainty handling, and reasoning within a single framework, improving both accuracy and interpretability without requiring additional supervision.

\section{Related Works}
\paragraph{\textbf{Gaze Target Prediction.}}
Early learning-based gaze methods relied on scene or activity priors that restricted where a target could be, which worked in constrained settings but struggled in open-world scenes. A shift to direct coordinate or heatmap prediction removed such assumptions and encouraged designs that combine person-specific and scene-level cues. The seminal two-branch framework of Recasens et al. on GazeFollow \cite{NIPS2015_ec895663} multiplied a viewer-independent scene saliency with a head-conditioned gaze mask to yield person-specific maps, and motivated many extensions on data and fusion \cite{Chong_2018_ECCV,Tafasca_2023_ICCV,Tafasca_2024_CVPR,JIN2022104924,Chong_2020_CVPR,zhao2020learning,Recasens_2017_ICCV,9828503,Zhang_2017}. Building on this, multi-branch models introduced additional modalities to reduce ambiguity and improve generalization. Depth was added either via monocular prediction or sensors so that near objects in the image but far in 3D could be separated; representative methods reconstruct point clouds or derive geometric cues such as front-most surfaces and angular offsets that guide head-conditioned decoding \cite{bao2022escnet,fang2021dual,Miao_2023_WACV,Tonini_2022,horanyi2023they,10.1145/3689643}. Pose-aware designs complemented head appearance with body keypoints and depth, often with a human branch that predicts a 2D gaze vector and a differentiable cone prior, and a scene branch that encodes RGB, depth, and pose maps with attention-based fusion and modality dropout for robustness \cite{Gupta_2022_CVPR,ranftl2020robustmonoculardepthestimation,hrformer}. A related direction estimated 3D head orientation and combined a gaze cone with depth rebasing so that only geometrically consistent regions remain, which also supports in or out of frame decisions \cite{horanyi2023they,fang2021dual}. To better handle extreme head poses or partial occlusion, face plus left and right eye streams preserved fine ocular detail before regressing pitch and yaw; attention and transformer-based fusion adaptively weighted facial versus ocular evidence \cite{GazeSymCAT,Chong_2018_ECCV,9050633,cai2021gazeestimationensemblearchitectures,10.1007/978-3-030-20876-9_20,10.1093/jcde/qwad038,5959981,Fischer_2018_ECCV}. Object-aware variants first detected heads and objects, then restricted reasoning to items that fall inside a fixed-angle field of view and biased transformer attention using cone to object alignment scores, which improved heatmaps and out-of-frame classification in clutter \cite{9828503,Tonini_2023_ICCV,Zhou_2024_CVPR,Wang_Guo_Jin_Xia_Liu_2024}.

\paragraph{\textbf{End-to-End and 3D Gaze Estimation.}}
In parallel, alternative formulations simplified or unified the pipeline. DETR-style set prediction removed upstream head detectors by jointly predicting a fixed-size set of human and gaze instances including head box, in or out decision, and heatmap, trained with Hungarian matching across boxes, classification, and regression \cite{Tonini_2023_ICCV,9879533}. A distinct track regressed 3D gaze direction without identifying a target, which is attractive for AR or driver monitoring but provides less semantics; these methods align 3D context in an egocentric frame and encode direction and distance relations among pose and objects with a transformer to refine the vector \cite{Jindal_2024_CVPR,Fan_2019_ICCV,10.1007/978-3-030-20893-6_3,Vuillecard_2025_CVPR,Nonaka_2022_CVPR,9740573,Kawana_2025_CVPR,Qin_2026_WACV}. To improve cross-domain robustness, contrastive approaches shaped feature geometry so that samples with similar gaze semantics align while identity or quality factors are suppressed; recent methods combine appearance-aware regression with language-driven differential contrast or distill toward vision–language spaces \cite{10.1145/3746027.3755096,du2023unsupervisedgazeawarecontrastivelearning,XIA2025111244,Wang_2022_CVPR,jindal2022contrastiverepresentationlearninggaze,yin2024clip}.

\paragraph{\textbf{Social and Multi-Person Gaze.}}
Researchers then moved beyond single-person localization to semantic and social gaze. Mutual gaze or looking at each other was recognized by fusing temporal head pose with spatial context \cite{doosti2021boosting,Marin-Jimenez_2019_CVPR}. Joint attention estimated a shared focus using interaction-aware transformers over per-person attributes together with scene-based attention maps \cite{Nakatani_2023_ICCV,Chong_2020_CVPR}. Multi-person temporal frameworks went further by producing per-person heatmaps, in or out predictions, and pairwise social labels such as looking at head, looking at each other, and shared attention through people–scene and spatio–temporal interaction modules \cite{NEURIPS2024_1caf09c9}. Unified token-based encoders achieved compact inference by fusing gaze tokens derived from head crops and bounding boxes with image tokens and decoding heatmaps and in or out labels in a single pass \cite{Tafasca_2024_CVPR}.

\paragraph{\textbf{Vision Language Models for Gaze.}}
With the rise of vision language models \cite{radford2021learning, li2022blip, li2023blip, NEURIPS2023_6dcf277e, team2023gemini, NEURIPS2024_dc06d4d2, molmo, bai2025qwen25vltechnicalreport, Qwenvl3,an2025llava}, gaze modeling has begun to benefit from open-vocabulary and zero-shot capabilities. These models enable richer multimodal reasoning by aligning visual representations with natural language, allowing systems to interpret attention in a more flexible and semantically meaningful way. As a result, gaze prediction can move beyond predefined object categories and instead relate gaze targets to descriptive concepts expressed in language.

\paragraph{\textbf{Semantic Gaze Understanding.}}
Several works have attempted to integrate gaze estimation with broader visual understanding pipelines. Some approaches combine gaze transformers with image–text matching or object detection frameworks to associate predicted gaze with candidate objects in the scene. Other methods adopt unified architectures that jointly predict both the gaze location and the identity of the attended object within a single model \cite{tu2023jointgazelocationgazeobjectdetection,NEURIPS2024_dbeb7e62}. More recently, GazeLLM \cite{gazellm} introduces a zero-shot reasoning framework that leverages large language models to infer gaze targets through spatial–semantic reasoning over structured scene representations.
Despite these advances, existing approaches remain constrained to predefined object categories or depend on intermediate object detection pipelines, limiting their ability to leverage open-vocabulary reasoning. As a result, several challenges in multimodal gaze-based scene understanding remain unresolved.

\section{Architecture}
\begin{figure}[t]
    \begin{adjustwidth}{-1cm}{-1cm} 
    \centering
    \includegraphics[width=1.075\textwidth]{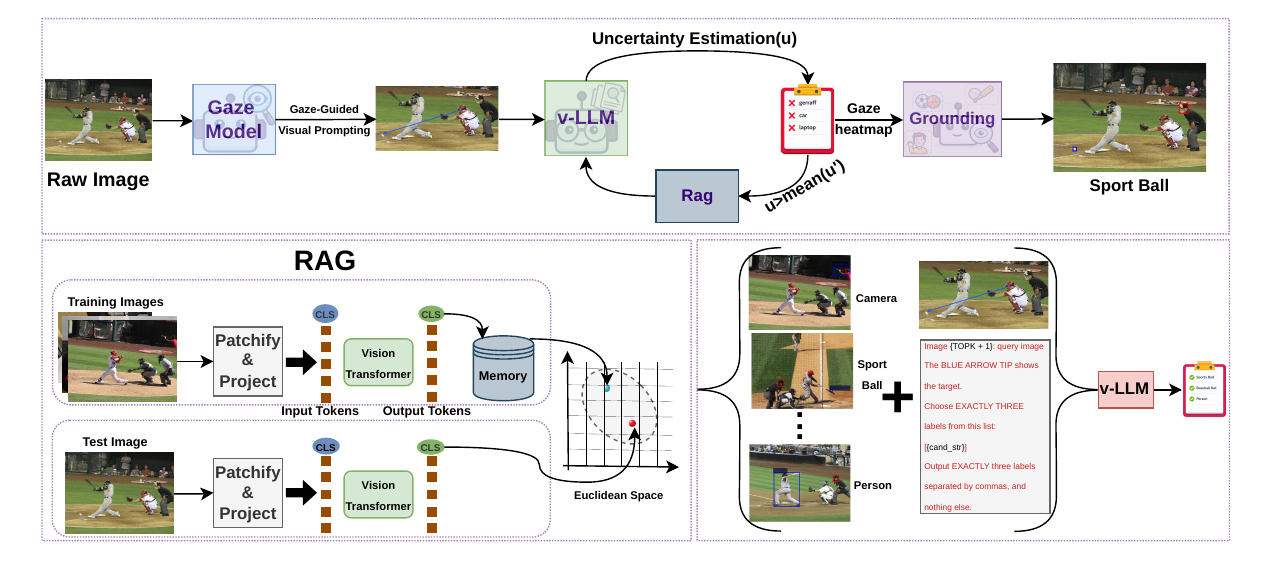}
    \end{adjustwidth}
    \caption{{Overview of the proposed gaze-target reasoning agent.}
    Given a raw image, the agent first detects head bounding boxes and estimates the gaze direction of each person to infer where they are looking. The predicted gaze is converted into a visual prompt that highlights the attended region and is provided to a vision language model (VLM). The VLM predicts the object being observed and outputs an uncertainty score. When the prediction uncertainty is high, a retrieval augmented generation (RAG) module retrieves visually similar images and their labels from a memory bank and feeds them, together with the query image, back to the VLM to improve reasoning. Finally, a grounding module localizes the predicted object in the image.
    }
    \label{fig:framework}
    
\end{figure}

Our goal is to develop a fully training-free agent that leverages human gaze as a guiding signal for attention-aware and open-vocabulary reasoning. As illustrated in Figure~\ref{fig:framework}, the proposed framework integrates pretrained components into a unified pipeline that jointly performs gaze estimation, object reasoning, and grounding. Given an input image, the agent first detects individuals and estimates their gaze, which serves as an explicit attentional cue for subsequent reasoning. These gaze signals guide a vision–language model to identify attended objects and interpret scene interaction, while a grounding module localizes relevant regions in the image.
To further support reasoning without additional training, the framework incorporates a lightweight memory mechanism that stores visual representations from the training data. During inference, the agent evaluates the uncertainty of its predictions, and when the uncertainty exceeds a predefined threshold, the system retrieves the most relevant examples from this memory to provide contextual guidance for interpreting the scene. This retrieval-based refinement helps the model resolve ambiguous gaze predictions by leveraging similar visual contexts.

\subsection{Gaze-Guided Visual Encoding}

The first stage of our gaze target agent(GTA) extracts person-specific attention signals from the input image. Given an RGB image, we begin by detecting all visible heads using a pretrained YOLOv8 detector~\cite{yaseen2024yolov8indepthexplorationinternal} without additional fine-tuning. Accurate head localization is essential, as gaze estimation is conditioned on the detected head regions. We adopt a head detector rather than a face detector because it is more robust to profile views, back views, and partial occlusions, enabling more consistent subject identification in unconstrained multi-person scenes.
For each detected head, we estimate gaze using the zero-shot Gaze-LLE model~\cite{Ryan_2025_CVPR}. A frozen DINOv2 encoder~\cite{dinov2} processes the image to produce a shared scene representation that is reused for all individuals, improving efficiency in crowded scenes. The head bounding box is converted into a binary mask and passed to a lightweight decoder composed of ViT blocks~\cite{vaswani2017attention}, which attends to the shared features and predicts both a fixation heatmap and an in-frame or out-of-frame decision. When the predicted fixation lies within the image, the heatmap is reduced to a single coordinate representing the estimated point of gaze.

To integrate gaze into subsequent reasoning without modifying the vision language model, we convert each predicted fixation into an explicit visual cue. For every detected individual, we draw a blue arrow from the center of the head box toward the predicted fixation location. If the gaze is predicted to lie outside the frame, the arrow extends toward the nearest image boundary to indicate outward attention. Given an image containing $N$ individuals, we generate $N$ prompted images $\{I'_i\}_{i=1}^{N}$, where $I'_i$ contains only the gaze arrow corresponding to person $i$. This allows the model to reason about each individual's gaze target independently.
Unlike traditional gaze-following approaches that represent attention as heatmaps or coordinates alone~\cite{NIPS2015_ec895663,Chong_2020_CVPR}, our framework explicitly encodes gaze as structured geometric cues that can be directly interpreted by modern vision language models~\cite{li2023blip,NEURIPS2023_6dcf277e,team2023gemini,Qwenvl3,an2025llava}. This design preserves interpretability while enabling gaze-aware reasoning without additional training or architectural modification.

\subsection{Vision Language Reasoning and Retrieval}

The final stage translates gaze-conditioned visual inputs into structured language outputs. For each prompted image, the vision language model (VLM) generates three candidate textual predictions together with their corresponding logits. These logits allow us to estimate the confidence of each prediction rather than relying solely on the decoded text.
At each decoding step $t$, the model produces logits $\mathbf{z}_t \in \mathbb{R}^{|V_{\text{VLM}}|}$ over the vocabulary $V_{\text{VLM}}$. These logits are converted into a probability distribution using softmax:\begin{equation}
P(w_t = w \mid \mathbf{x},\, w_{1:t-1})
=
\frac{\exp\!\left(z_t^{(w)}\right)}
{\sum_{w' \in V_{\text{VLM}}} \exp\!\left(z_t^{(w')}\right)}.
\end{equation}

We compute the probability of the first generated token $\hat{w}_1$:
\begin{equation}
    p_{\text{first}}
    \;=\;
    P(\hat{w}_1 \mid \mathbf{x}),
\end{equation}

The uncertainty score is then defined as $u = 1 - p_{\text{first}}$, where a lower value of $p_{\text{first}}$ indicates reduced model confidence in its initial prediction. Retrieval refinement is activated when the uncertainty exceeds the mean uncertainty $\mathbb{E}[u]$ computed over the model's previous predictions.
For the training images, we compute global representations using a pretrained Vision Transformer (ViT) encoder \cite{radford2021learning}, extracting the final-layer \texttt{[CLS]} token and storing these embeddings in a memory bank.
When a test sample is flagged as uncertain, the GTA activates a retrieval-based reasoning module. The gaze-conditioned test image is encoded using the same ViT encoder to obtain a query embedding. The GTA then compares this query embedding with feature embeddings stored in an external memory, which contains embeddings of gaze-conditioned training images together with their corresponding object labels. Similarity is measured by computing the Euclidean distance between the query embedding and the stored training embeddings, allowing the system to retrieve visually similar examples from the memory bank.

Based on this similarity measure, the agent retrieves the six nearest neighbors from the memory bank. Each retrieved item consists of a gaze-conditioned image and its associated object label. These examples are appended to the vision–language model prompt as in-context demonstrations, allowing the model to leverage similar previously observed attention patterns when resolving ambiguous cases. This retrieval-based conditioning enables the system to refine uncertain predictions without updating any model parameters, preserving the training-free nature of the gaze target agent.
Through this uncertainty-aware and retrieval-augmented reasoning process, the agent combines gaze estimation, confidence estimation, and external memory to refine uncertain gaze-target predictions without updating any model parameters.

\subsection{Gaze-Guided Object Grounding}

While gaze estimation provides a fixation point indicating the direction of attention, it does not precisely localize the full spatial extent of the attended object. In cluttered scenes where multiple similar objects appear close together, a gaze prediction may indicate the correct region but not the exact instance. This is particularly important for shared attention analysis, as predicting the object category alone does not guarantee that multiple individuals are attending to the same physical object. Accurate instance level grounding is therefore necessary to ensure spatial consistency in social gaze reasoning.
To localize the attended object, we incorporate a dedicated object grounding stage that generates candidate object regions in the scene. We evaluate several pretrained detection and segmentation-based grounding models, including RF-DETR~\cite{RF-DET}, YOLO26~\cite{sapkota2026yolo26keyarchitecturalenhancements}, and SAM3~\cite{carion2025sam}. These models produce candidate object regions in the image, either as bounding boxes or segmentation masks. Given the predicted gaze heatmap, we measure its spatial overlap with each candidate region and select the region with the highest interaction with the gaze distribution. This gaze-heatmap alignment enables the system to identify the object instance most consistent with the subject's visual attention, particularly in scenes containing multiple nearby objects.
Table~\ref{tab:gazehoi_results} reports the grounding performance of RF-DETR within our gaze-guided framework. Additional comparisons with other grounding models are provided in Appendix A.1 of the supplementary material.

\section{Experiments}
In this section, we evaluate the proposed \textbf{G}aze \textbf{T}arget \textbf{A}gent \textbf{(GTA)}, which combines gaze-guided visual prompting with uncertainty aware reasoning and memory based retrieval. To analyze the contribution of the retrieval and uncertainty components, we ablate on a simplified variant referred to as \textbf{G}aze \textbf{G}uided \textbf{V}ision \textbf{L}anguage \textbf{(GGVL)}\cite{nasiriboukani2025gaze}. This variant removes the uncertainty estimation and memory-based retrieval modules and relies solely on gaze conditioned visual prompts to guide the vision language model in predicting the attended object. Since localization performance is determined by the gaze estimator, we focus our analysis on the recognition capability of the proposed agent. Additionally, we compare GTA with state-of-the-art gaze target detection method.

\label{sec:experiments}

\begin{table*}[t]
\centering
\small
\setlength{\tabcolsep}{2.5pt}
\caption{\textbf{Performance comparison on \textit{GazeFollow} and GazeHOI.} 
We compare GGVL and the proposed GTA against the previous state-of-the-art (SOTA).
For \textit{GazeFollow}, we report Acc@1, Acc@3, and MultiAcc@1, while for GazeHOI we report Acc@1, Acc@3, AP@50, and mIoU. 
Grey marks the previous SOTA, while blue highlights GTA. 
Here, I and V denote image and vocabulary, respectively.}
\label{tab:GazeFollow_gazehoi_results}

\begin{subtable}[t]{0.496\textwidth}
\renewcommand{\arraystretch}{1.25}
\centering
\caption{\textbf{\textit{GazeFollow}.}Params denotes the number of learnable parameters.}
\label{tab:GazeFollow_results}
\resizebox{\linewidth}{!}{
\begin{tabular}{lccccc}
\toprule
\textbf{Method} 
& \textbf{Params} 
& \textbf{Input} 
& \textbf{Acc@1} $\uparrow$ 
& \textbf{Acc@3} $\uparrow$ 
& \textbf{MultiAcc@1} $\uparrow$ \\
\midrule
\rowcolor{gray!10}
Tafasca et al.\cite{NEURIPS2024_dbeb7e62} 
& 116M & I+V & 0.447 & 0.642 & \underline{0.516} \\

\midrule
Qwen3\textsubscript{[2B (GGVL)]} 
& 0 & I & \underline{0.450} & \underline{0.684} & 0.506 \\

\rowcolor{blue!10}
\textbf{Qwen3}\textsubscript{[2B (GTA)]} 
& 0 & I & \textbf{0.493} & \textbf{0.728} & \textbf{0.558} \\
\bottomrule
\end{tabular}
}
\end{subtable}
\hfill
\begin{subtable}[t]{0.496\textwidth}
\centering
\caption{\textbf{GazeHOI.}The $\dagger$ symbol indicates the previous state-of-the-art training-free result.}
\label{tab:gazehoi_results}
\resizebox{\linewidth}{!}{
\begin{tabular}{lccccc}
\toprule
\textbf{Method} 
& \textbf{GazeAcc} $\uparrow$ 
& \textbf{Acc@1} $\uparrow$
& \textbf{Acc@3} $\uparrow$
& \textbf{AP@50} $\uparrow$
& \textbf{mIoU} $\uparrow$ \\
\midrule
\rowcolor{gray!10}
Tafasca et al.\textsuperscript{$\dagger$}\cite{NEURIPS2024_dbeb7e62} 
& 0.652 & 0.306 & 0.463 & -- & -- \\

Tafasca et al.\cite{NEURIPS2024_dbeb7e62} 
& 0.723 & 0.583 & 0.706 & -- & -- \\

\midrule
Qwen3\textsubscript{[2B (GGVL)]} 
& \textbf{0.731} & \underline{0.517} & \underline{0.721} & \textbf{0.192} & \textbf{0.408} \\

\rowcolor{blue!10}
\textbf{Qwen3}\textsubscript{[2B (GTA)]} 
& \textbf{0.731} & \textbf{0.545} & \textbf{0.745} & \textbf{0.192} & \textbf{0.408} \\
\bottomrule
\end{tabular}
}
\end{subtable}

\end{table*}

\subsection{Quantitative Evaluation}
We evaluate recognition performance using three metrics: Acc@1, Acc@3, and MultiAcc@1. Acc@1 and Acc@3 measure whether the correct object label appears within the top 1 or top 3 predicted categories, respectively. MultiAcc@1 accounts for ambiguous cases where multiple object categories may plausibly correspond to the same gaze target, in such cases, a prediction is considered correct if the top 1 prediction matches any of the plausible object labels associated with the gaze region.
\paragraph{\textbf{GazeFollow}}
For fair comparison with prior work, we adopt the same target vocabulary as Tafasca et al.~\cite{NEURIPS2024_dbeb7e62}, ensuring that performance differences are not influenced by discrepancies in the label space. Under this controlled setting, as shown in Table~\ref{tab:GazeFollow_results}, our baseline GGVL\cite{nasiriboukani2025gaze} already achieves a clear improvement over previous methods, increasing Top 1 accuracy by 0.3\% and Top 3 accuracy by 4.2\%. These gains highlight the effectiveness of directional prompting for semantic gaze recognition, which enables the vision language model to better align gaze cues with meaningful object categories.
Building upon this baseline, the gaze target agent further improves recognition performance across all reported metrics. Specifically, it achieves an additional gain of 4.3\% in Top 1 accuracy and 4.4\% in Top 3 accuracy compared to GGVL, while also improving multi-label accuracy by 5.2\%. This consistent improvement across metrics indicates that the proposed agent framework not only enhances single instance predictions but also performs reliably in scenes containing multiple interacting subjects.

\paragraph{\textbf{GazeHOI}}

We further evaluate our method on \textit{GazeHOI} (Table~\ref{tab:gazehoi_results}). In addition to recognition accuracy, we also report grounding performance on this dataset, measuring how well the predicted semantic labels align with the annotated human object interaction targets.
For object localization, we use the RF-DETR\cite{RF-DET} detector to generate candidate object bounding boxes and select the bounding box whose region overlaps most with the predicted gaze heatmap, treating it as the predicted gaze target object.

Even when compared against fine-tuned methods, our approach remains competitive. Notably, GTA achieves a 3.9\% higher Top 3 accuracy than the baseline, further demonstrating the effectiveness of our design. These results highlight the effectiveness of the GTA framework as a unified zero-shot system. The gains mainly come from stronger semantic reasoning over the gaze prior, rather than task-specific training or changes to the localization mechanism.

\subsection{Ablation Study} 
We conduct ablation studies to analyze the main components of the proposed gaze target agent. We first evaluate whether GTA generalizes across different VLM backbones and model scales. For the initial ablation setting, we use a blue arrow visual prompt with a width of $4$ px, retrieve the top $6$ nearest examples, and apply dynamic uncertainty thresholding. We then study uncertainty threshold, visual prompt design, retrieval size, open-vocabulary recognition, and computational cost. Except for the backbone and scale analysis, all ablations are conducted using Qwen3-VL 2B on the GazeFollow dataset. The best-performing configuration identified from these ablations is used for the final GTA result reported in Table~\ref{tab:GazeFollow_results}.

\begin{table*}[t]
\centering
\caption{Ablation study on \textit{GazeFollow} and GazeHOI comparing GGVL and the proposed GTA across different public and private VLM backbones. Results are reported under identical settings to show the effect of reasoning across model scales and architectures.}
\label{tab:scales_ablation}
\begin{threeparttable}
\resizebox{\textwidth}{!}{%
\renewcommand{\arraystretch}{1.00}
\begin{tabular}{l cc ccc}
\toprule
\multirow{2}{*}{\textbf{Method}} &
\multicolumn{2}{c}{\textbf{GazeHOI}} &
\multicolumn{3}{c}{\textbf{GazeFollow}} \\
\cmidrule(lr){2-3}\cmidrule(lr){4-6}
& \textbf{Acc@1} $\uparrow$ & \textbf{Acc@3} $\uparrow$ &
\textbf{Acc@1} $\uparrow$ & \textbf{Acc@3} $\uparrow$ & \textbf{MultiAcc@1} $\uparrow$ \\
\midrule
\rowcolor{gray!10}
\multicolumn{6}{l}{\textbf{Public}} \\
\midrule
Qwen3\textsubscript{[2b (GGVL)]} & 0.517 & 0.721 & 0.450 & 0.684 & 0.506 \\
\rowcolor{blue!10}
\textbf{Qwen3\textsubscript{[2b (GTA)]}} & \textbf{0.545} & \textbf{0.745} & \textbf{0.488} & \textbf{0.717} & \textbf{0.550} \\
\midrule
Llava 1.5\textsubscript{[4b (GGVL)]} & 0.528 & 0.691 & 0.380 & 0.630 & 0.437 \\
Llava 1.5\textsubscript{[4b (GTA)]} & \textbf{0.572} & 0.750 & 0.466 & 0.700 & 0.535 \\
Qwen3\textsubscript{[4b (GGVL)]} & 0.516 & 0.742 & 0.543 & 0.722 & 0.613 \\
\rowcolor{blue!10}
\textbf{Qwen3\textsubscript{[4b (GTA)]}} & \underline{0.564} & \textbf{0.758} & \textbf{0.559} & \textbf{0.769} & \textbf{0.636} \\
\midrule
Llava\textsubscript{[8b (GGVL)]} & 0.548 & 0.718 & 0.409 & 0.676 & 0.468 \\
Llava\textsubscript{[8b (GTA)]} & 0.557 & 0.771 & 0.484 & 0.726 & 0.553 \\
Qwen3\textsubscript{[8b (GGVL)]} & 0.533 & 0.742 & 0.584 & 0.755 & 0.650 \\
\rowcolor{blue!10}
\textbf{Qwen3\textsubscript{[8b (GTA)]}} & \textbf{0.602} & \textbf{0.772} & \textbf{0.602} & \textbf{0.778} & \textbf{0.668} \\
\midrule
\rowcolor{gray!10}
\multicolumn{6}{l}{\textbf{Private}} \\
\midrule
Gemini 2.5 Flash\textsubscript{[GGVL]} & 0.617 & 0.741 & 0.594 & 0.773 & 0.655 \\
\rowcolor{blue!10}
\textbf{Gemini 2.5 Flash\textsubscript{[GTA]}} & \textbf{0.621} & \textbf{0.753} & \textbf{0.622} & \textbf{0.789} & \textbf{0.679} \\
Gemini 2.5 pro\textsubscript{[GGVL]} & 0.625 & 0.728 & 0.584 & 0.767 & 0.649 \\
Gemini 2.5 pro\textsubscript{[GTA]} & 0.631 & 0.739 & 0.608 & 0.787 & 0.665 \\
\bottomrule
\end{tabular}
}
\end{threeparttable}
\end{table*}

\paragraph{\textbf{Generalization Across VLM Backbones and Model Scales.}}

To assess whether the proposed agent-based reasoning is robust across different model families and capacities, we compare the baseline GGVL framework with GTA under identical settings, as reported in Table~\ref{tab:scales_ablation}. In this experiment, the gaze localization module remains fixed, allowing us to isolate the effect of the recognition and reasoning components. Our evaluation includes both public and private VLMs. We evaluate both public VLMs, including LLaVA and Qwen3 at different scales, and private VLMs, including Gemini 2.5 Flash and Pro.

Across all public VLMs and parameter sizes, GTA consistently improves over the GGVL baseline on both GazeFollow and GazeHOI. The gains are stable from smaller to larger models, indicating that the benefits of structured reasoning are not dependent on model scale. Notably, LLaVA shows a particularly strong boost when moving from GGVL to GTA in both the 4B and 8B configurations, with clear improvements across Top 1, Top 3, and multi-label accuracy metrics.
Gemini results further show that the gains are not limited to open-source models. Overall, the ablation study shows that the observed performance gains are consistent across architectures, parameter scales, and both public and private VLMs, validating the generality of the proposed gaze target agent.

\paragraph{\textbf{Uncertainty Threshold}}

The results in Table~\ref{tab:uncertainty_threshold} show that, excluding the retrieve-all setting at threshold $0$, performance remains stable across different uncertainty thresholds. The weaker performance of the retrieve-all setting suggests that applying retrieval indiscriminately can introduce noisy or confusable candidates for samples that the model already predicts correctly and confidently, leading the model away from its correct prediction. The dynamic threshold achieves the best top 1 and multi-label accuracy, while the fixed threshold of $25$ gives the best top 3 and closely matches the dynamic setting across all metrics. This is because the dynamic threshold converges to an optimal value close to $25$ after around $20$--$30$ steps, which explains the similar performance of the fixed $25$ threshold.

\paragraph{\textbf{Different Visual Prompt Designs.}} Different visual prompt designs have a clear impact on gaze-target recognition, as shown in Table~\ref{tab:visual_prompt}. The head bounding box performs poorly with only $28.69$ top-1 accuracy, indicating that head location alone is insufficient without gaze direction. Non-arrow prompts such as the blue dot, grayscale masking, and blur masking improve performance, but remain below arrow-based prompts. This shows that explicitly encoding gaze direction is more effective. Among single-arrow prompts, the blue arrow achieves the best result with $48.80$ top-1 accuracy, slightly outperforming the red arrow, while the green arrow performs lower. We further find that arrow width has a smaller but noticeable effect: a $5$px blue arrow achieves the best overall performance with $49.29$ top-1 accuracy and $55.79$ multi-label accuracy. Overall, the $5$px blue arrow provides the best balance between visibility and limited occlusion.

\paragraph{\textbf{Retrieval-Size Ablations}}

The retrieval-size ablation in Table~\ref{tab:retrieval_size} shows that $k=6$ gives the best overall performance. Retrieval helps the model in two ways: it provides relevant examples for uncertain cases and reduces the effective search space from the full vocabulary of more than 300 fixed labels. This makes the reasoning process more focused and prevents the VLM from being overwhelmed by many possible object categories. Smaller values of $k$ provide less context, while larger values may introduce less relevant examples and slightly reduce performance.

\begin{table}[t]
\centering
\begin{threeparttable}
\caption{\textbf{Uncertainty threshold ablation.} We compare fixed thresholds with the dynamic uncertainty threshold used in GTA.}
\label{tab:uncertainty_threshold}
\renewcommand{\arraystretch}{1}
\setlength{\tabcolsep}{8pt}
\begin{tabular}{c c c c}
\toprule
Threshold & Acc@1 $\uparrow$ & Acc@3 $\uparrow$ & MultiAcc@1 $\uparrow$ \\
\midrule
0  & 42.64 & 71.25 & 49.23 \\
15 & 47.85 & \underline{71.77} & 54.33 \\
25 & \underline{48.49} & \textbf{71.83} & \underline{54.77} \\
35 & 48.47 & 71.54 & 54.75 \\
45 & 47.76 & 70.85 & 53.97 \\
55 & 47.09 & 70.03 & 53.09 \\
\rowcolor{blue!10}
Dynamic & \textbf{48.80} & 71.70 & \textbf{55.50} \\
\bottomrule
\end{tabular}
\end{threeparttable}
\end{table}

\paragraph{\textbf{Open-Vocabulary Evaluation.}}

We evaluate GTA in both fixed and open vocabulary settings. The open-vocabulary setting removes the predefined label constraint and evaluates whether the model can generate meaningful free-form predictions.
As shown in Table~\ref{tab:open_vocab_ablation}, GTA improves performance in both fixed and open-vocabulary settings, with larger gains in the open setting. In the fixed setting, GTA improves Top-1 by $3.54$ points, Top-3 by $3.03$ points, and MultiAcc@1 by $4.31$ points. In contrast, in the open-vocabulary setting, GTA improves Top-1 by $9.79$ points, Top-3 by $10.52$ points, and MultiAcc@1 by $10.48$ points. This shows that agent-based reasoning and retrieval refinement are especially effective when predictions are not restricted to a predefined label set.

Exact string matching can underestimate open-vocabulary performance because valid semantic variants may not exactly match the ground-truth label. For example, predictions such as ``phone'' and ``cell phone'' may refer to the same object, while labels such as ``person'' may be described as ``man,'' ``woman,'' ``player,'' or ``face.'' We also report E5~\cite{wang2022text} semantic matching, where predictions are considered correct if their cosine similarity with the ground-truth label is at least $0.90$. Open + GTA achieves $47.85$ Top-1, while Fixed + GTA reaches $51.48$, confirming that GTA produces semantically closer gaze-target predictions.

\begin{table}[t]
\centering
\caption{\textbf{Visual prompt ablation.} We compare different prompt types, arrow colors, and arrow widths. The color arrows without a stated width use $4$ px.}
\label{tab:visual_prompt}
\centering
\resizebox{0.9\linewidth}{!}{
\begin{tabular}{p{4cm} c c c}
\toprule
\textbf{Prompt} & \textbf{~~~~Acc@1} $\uparrow$ ~~~~& \textbf{~~~~Acc@3} $\uparrow$~~~~ & \textbf{~~~~MultiAcc@1}$\uparrow$~~~~ \\
\midrule
Head bbox & 28.69 & 67.57 & 34.36 \\
Blue dot & 40.99 & 66.90 & 47.49 \\
Gray scale & 38.39 & 69.72 & 44.42 \\
Blur & 38.98 & 67.75 & 44.92 \\
Red arrow & 48.29 & 72.08 & 54.45 \\
Green arrow & 43.91 & 71.98 & 50.17 \\
Yellow arrow & 48.06 & 72.21 & 54.25 \\
Blue arrow & 48.80 & 71.70 & \textbf{55.50} \\
3px blue arrow & 48.41 & 72.25 & 54.83 \\
\rowcolor{blue!10}
\textbf{5px blue arrow} & \textbf{49.29} & \underline{72.81} & \textbf{55.79} \\
6px blue arrow & \underline{49.10} & \textbf{72.84} & \underline{55.40} \\
\bottomrule
\end{tabular}
}
\end{table}

\begin{table}[t]
\centering
\begin{threeparttable}
\caption{\textbf{Retrieval-size ablation.} We vary the number of retrieved examples used during retrieval-based refinement.}
\label{tab:retrieval_size}
\renewcommand{\arraystretch}{1}
\setlength{\tabcolsep}{10pt}
\begin{tabular}{c c c c}
\toprule
$k$ & Acc@1 $\uparrow$ & Acc@3 $\uparrow$ & MultiAcc@1 $\uparrow$ \\
\midrule
3  & 48.68 & \underline{72.75} & 55.02 \\
4  & 48.56 & 72.69 & 54.89 \\
5  & \underline{49.21} & \textbf{72.81} & \underline{55.58} \\
\rowcolor{blue!10}
6  & \textbf{49.29} & \textbf{72.81} & \textbf{55.79} \\
10 & 49.04 & 72.27 & 55.37 \\
\bottomrule
\end{tabular}
\end{threeparttable}
\end{table}

\paragraph{\textbf{Computational Cost.}}
We also analyze the computational cost of the proposed retrieval-based refinement. The GGVL baseline, which directly sends the gaze-guided visual prompt to the VLM, requires approximately $4.94{\times}10^{12}$ FLOPs per sample. When uncertainty exceeds the threshold, GTA additionally activates retrieval refinement, adding approximately $9.73{\times}10^{12}$ FLOPs for that sample. With the dynamic threshold, retrieval is triggered for about $45\%$ of samples, which increases the average cost. A more efficient alternative is to use a fixed threshold of $55$, which still preserves most of the performance gain while triggering retrieval for only $729/4781$ samples, or approximately $15.2\%$ of the test set. In this setting, the average cost becomes
\[
(1-0.152)\,4.94{\times}10^{12}
+
0.152\,(4.94+9.73){\times}10^{12}
\approx
6.42{\times}10^{12}
\]
FLOPs per sample. Among the $729$ triggered samples, $520$ were originally incorrect, showing that the uncertainty score effectively identifies challenging cases while avoiding unnecessary retrieval calls.

\begin{table}[t]
\caption{\textbf{Vocabulary-setting ablation.}
We compare fixed and open-vocabulary evaluation for GGVL and GTA using both exact matching and E5 semantic matching.}
\label{tab:open_vocab_ablation}
\centering
\resizebox{0.99\linewidth}{!}{
\begin{tabular}{l c c c c c}
\toprule
\textbf{Vocabulary} & ~~\textbf{Acc@1} $\uparrow$~~ & ~~\textbf{Acc@3} $\uparrow$~~ & ~~\textbf{MultiAcc@1} $\uparrow$ ~~
& ~~\textbf{E5 Acc@1} $\uparrow$ ~~& ~~\textbf{E5 Acc@3} $\uparrow$ ~~\\
\midrule
Fixed + GGVL
& \underline{45.75} & \underline{69.78} & \underline{51.48} & \underline{48.43} & \underline{72.40} \\
\rowcolor{blue!10}
\textbf{Fixed + GTA }
& \textbf{49.29} & \textbf{72.81} & \textbf{55.79} & \textbf{51.48} & \textbf{76.29} \\
Open + GGVL
& 30.84 & 43.75 & 35.59 & 40.11 & 57.28 \\
Open + GTA 
& 40.63 & 54.27 & 46.07 & 47.85 & 64.62 \\
\bottomrule
\end{tabular}
}
\end{table}

\subsection{Qualitative Evaluation}
\begin{figure}[t]
    \centering

    \begin{subfigure}[t]{0.85\textwidth}
        \centering
        \includegraphics[width=\linewidth]{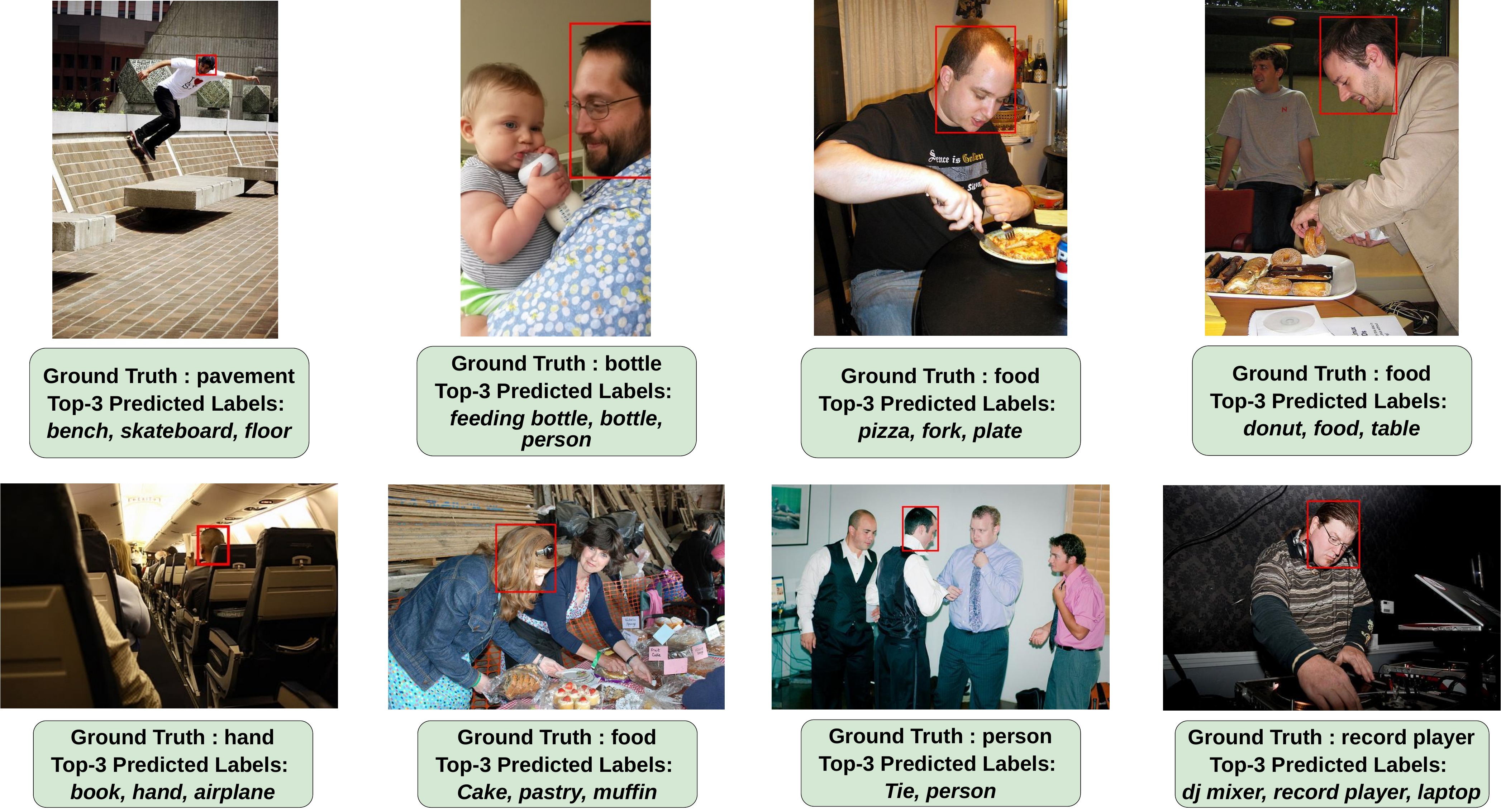}
        \caption{Examples of semantically richer gaze-target predictions generated by GTA.}
        \label{fig:semantic_predictions}
    \end{subfigure}

    \medskip

    \begin{subfigure}[t]{0.83\textwidth}
        \centering
        \includegraphics[width=\linewidth]{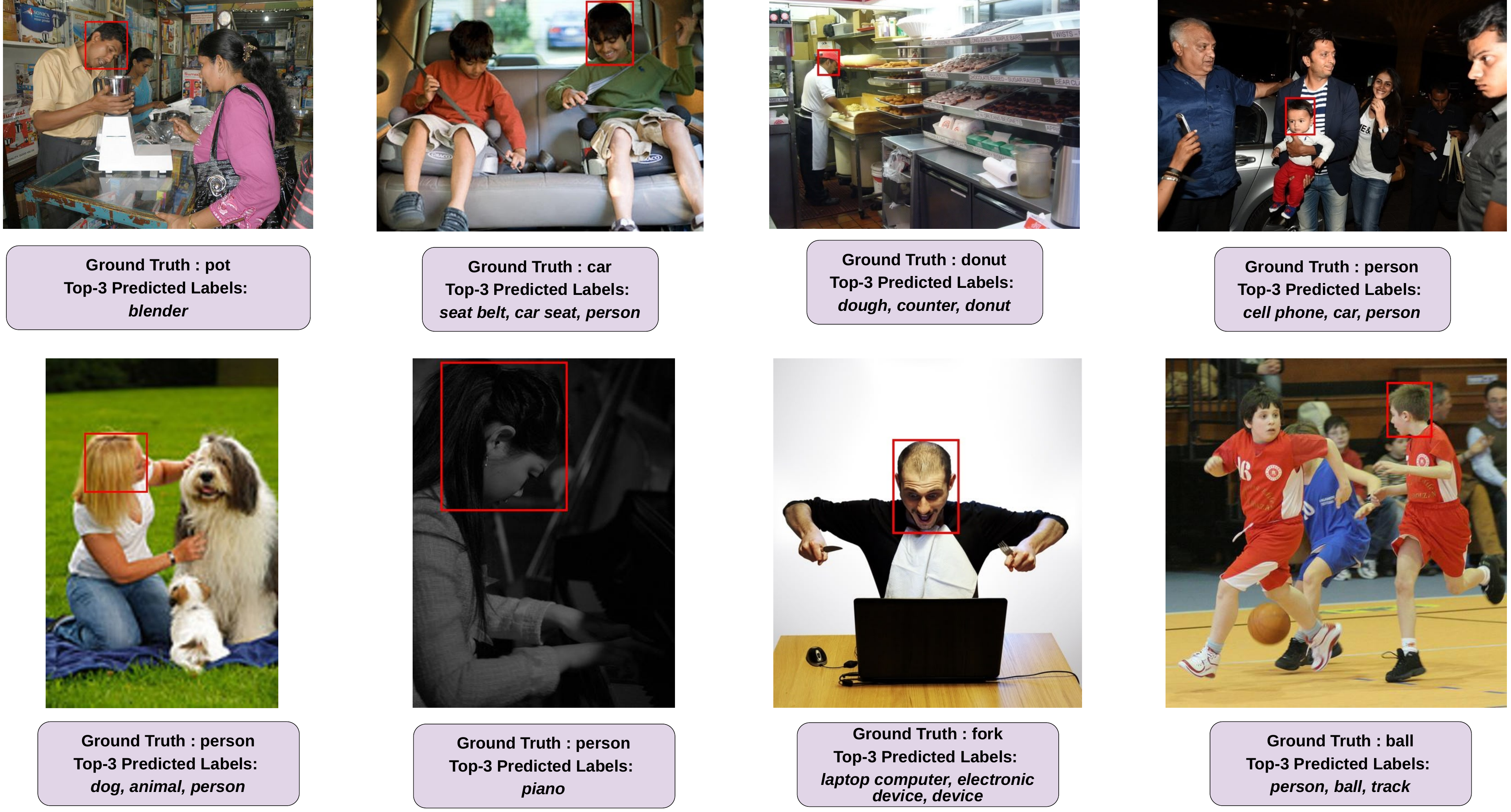}
        \caption{Examples with incorrect ground-truth labels, where GTA correctly predicts the attended object.}
        \label{fig:gt_errors}
    \end{subfigure}
    \label{fig:qualitative_examples}
    \caption{Examples illustrate GTA's semantic reasoning and robustness to annotation errors. The ground-truth label and top-3 model predictions are shown for each image.}
    \end{figure}
    
In addition to quantitative results, we conduct qualitative analyses to examine the reasoning capabilities of the proposed agent. The first two analyses use our agent with the predefined vocabulary, showing that it can produce richer semantic descriptions of attended objects and correctly identify gaze targets in cases with incorrect ground truth annotations. The final analysis removes the vocabulary constraint and demonstrates that the agent can generate flexible open-vocabulary predictions.

\paragraph{\textbf{Richer Semantic Predictions}}
Figure~\ref{fig:semantic_predictions} presents examples where the predicted labels provide more detailed semantic descriptions of the attended objects than the original dataset annotations. Rather than producing generic object categories, the agent often generates labels that better reflect the visual context of the scene. This demonstrates the ability of the framework to leverage the semantic knowledge of vision language models for more descriptive gaze target interpretations.

\paragraph{\textbf{Ground Truth Annotation Errors}}
Figure~\ref{fig:gt_errors} shows cases where the dataset ground truth annotations appear incorrect. In these examples, the agent predicts object labels that better correspond to the visual gaze target. These cases highlight limitations in existing annotations and illustrate how the model’s reasoning can reveal inconsistencies in the dataset.

\begin{figure}[t]
    \centering
    \includegraphics[width=0.9\textwidth]{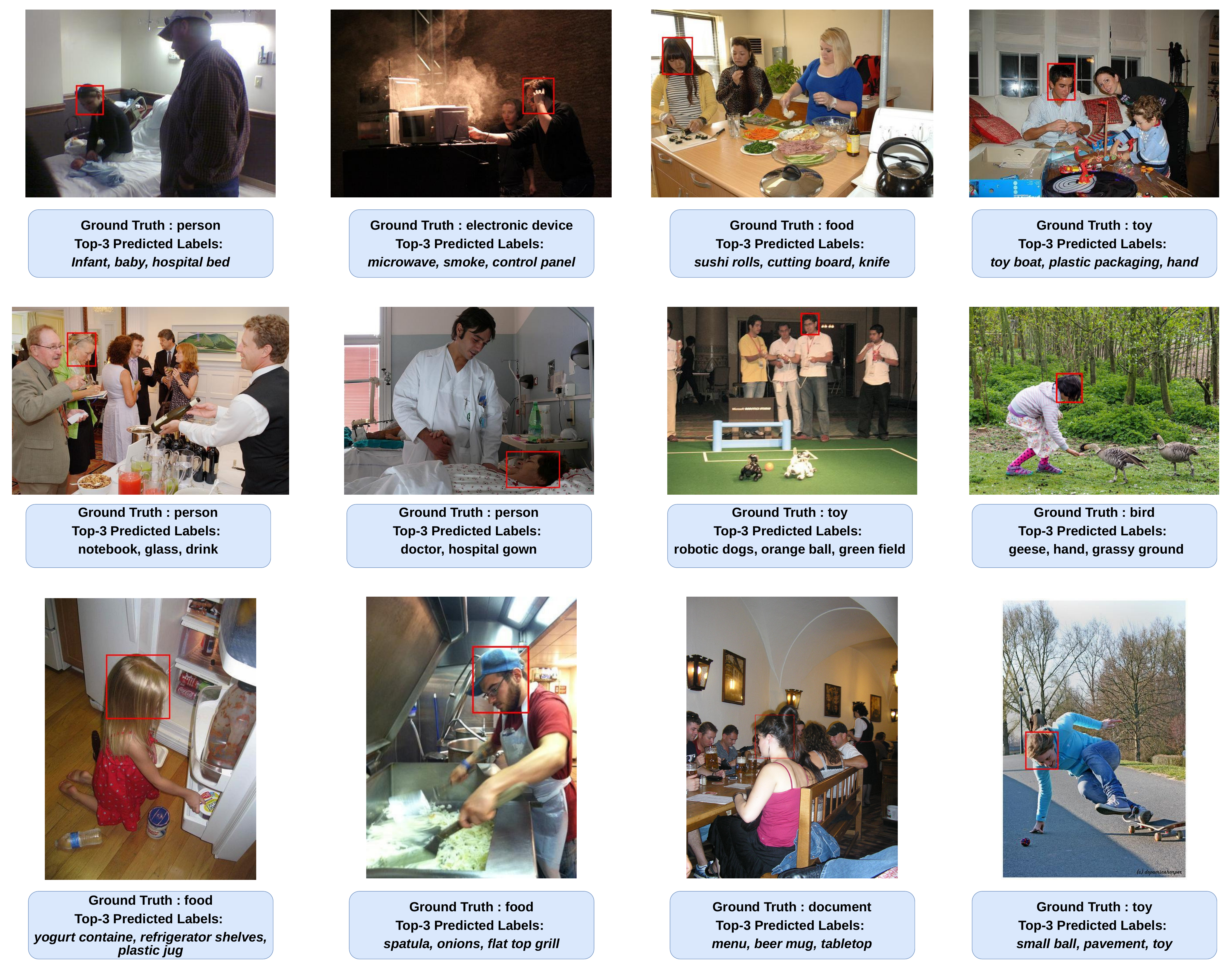}
    \caption{Examples where GTA is not restricted to a fixed vocabulary. The predicted labels frequently provide more precise descriptions than the ground truth annotations.}
    \label{fig:open_predictions}
\end{figure}

\paragraph{\textbf{Open-Vocabulary Predictions}}
Finally, Figure~\ref{fig:open_predictions} demonstrates the behavior of the agent when it is not restricted to a predefined vocabulary. Without label constraints, the model frequently generates more precise and context-aware descriptions of the attended objects. These results indicate that the agent is capable of leveraging open vocabulary reasoning to produce flexible and semantically rich predictions beyond the fixed categories used during quantitative evaluation.

\section{Conclusion}
\label{sec:conclusion}

We presented \textbf{GTA}, a fully training-free framework for gaze-guided semantic reasoning that unifies gaze localization, object recognition, grounding, and uncertainty-aware refinement within a single pipeline. By converting gaze predictions into explicit visual prompts, the approach enables off-the-shelf vision language models to reason about attended objects in a transparent and interpretable manner. A key component of our design is a lightweight memory module that is activated only for uncertain samples, retrieving semantically similar gaze-conditioned examples to refine predictions through in-context conditioning without any parameter updates.

Extensive experiments on \textit{GazeFollow} and \textit{GazeHOI} demonstrate that GTA achieves state of the art performance while remaining competitive with fine-tuned approaches. Qualitative analyses further show that GTA produces richer semantic interpretations of gaze targets, can recover correct predictions in the presence of annotation errors, and naturally supports open-vocabulary outputs when label constraints are removed. Together, these results indicate that gaze can serve as a powerful and general guiding signal for training-free visual reasoning, improving both accuracy and interpretability without additional supervision.

\section*{Acknowledgements}

This work was supported by the Engineering and Physical Sciences Research Council (EPSRC) through the University of Surrey Doctoral Landscape Award [grant number EP/Z535072/1 ].

The authors acknowledge the use of resources provided by the Isambard-AI National AI Research Resource (AIRR). Isambard-AI is operated by the University of Bristol and is funded by the UK Government’s Department for Science, Innovation and Technology (DSIT) via UK Research and Innovation; and the Science and Technology Facilities Council [ST/AIRR/I-A-I/1023]\cite{mcintoshsmith2024isambardaileadershipclasssupercomputer}.

\clearpage
\bibliographystyle{splncs04}
\bibliography{main}
\end{document}


\appendix
\section{Supplementary}
This supplementary material provides additional details and results for \textit{Gaze Target Agent}\textbf{(GTA)}. It includes extended detector comparisons, prompt templates used by the agent, and further qualitative examples illustrating the strengths and limitations of the method.

\subsection{Detector-Based Gaze-Guided Object Selection}

Table~\ref{tab:detector_comparison} provides an extended comparison of gaze-guided object selection strategies across multiple detectors on GazeHOI. For each detector, we report results under two proposal settings: using all detector outputs and restricting candidates to the top-3 predicted classes. We compare three selection modes: Upper Bound, which measures the best achievable match among the available proposals; Gaze-Point, which selects the proposal containing the peak gaze location; and Heatmap, which ranks proposals using the full spatial gaze distribution.

\begin{figure}[t]
    \centering
    \includegraphics[width=0.8
    \textwidth]{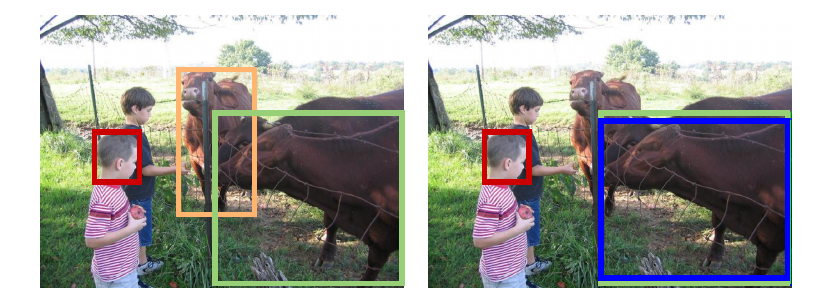}
    \caption{\textbf{Heatmap-based selection outperforms gaze-point selection.} 
    The green box is the ground-truth object, the red box is the head, the orange box is the gaze-point selection, and the blue box is the heatmap-based selection. Here, the heatmap aligns better with the target object, resulting in a more accurate selection.}
    \label{fig:heatmap_better}
\end{figure}

\begin{table*}[h]
\centering
\caption{Comparison of gaze-guided object selection across detectors on GazeHOI. For each detector, results are reported for two candidate sets: \emph{All Detections} and \emph{Top-3 Classes}. We compare three selection modes: \emph{Upper Bound}, \emph{Gaze-Point}, and \emph{Heatmap}.}
\small
\setlength{\tabcolsep}{5pt}
\begin{tabular}{lcccccc}
\toprule
& \multicolumn{3}{c}{All Detections} & \multicolumn{3}{c}{Top-3 Classes} \\
\cmidrule(lr){2-4} \cmidrule(lr){5-7}
Method & mIoU $\uparrow$ & R@50 $\uparrow$ & AP@50 $\uparrow$ & mIoU $\uparrow$ & R@50 $\uparrow$ & AP@50 $\uparrow$ \\
\midrule

\rowcolor{gray!10}
\multicolumn{7}{l}{\textbf{RF-DETR}} \\

Upper Bound
& 62.65 & 67.14 & 46.60 & 60.79 & 64.66 & 43.03 \\

Gaze-Point 
& 38.80 & 38.30 & 17.32 & 38.71 & 38.22 & 17.33 \\
\rowcolor{blue!10}
Heatmap
& 40.81 & 40.72 & 19.16 & 42.04 & 41.72 & 19.65 \\

\midrule

\rowcolor{gray!10}
\multicolumn{7}{l}{\textbf{Yolo26}} \\

Upper Bound & 56.23 & 59.58 & 36.88 & 44.69 & 45.96 & 24.58 \\
Gaze-Point & 37.37 & 36.98 & 15.10 & 31.20 & 30.22 & 11.22 \\
\rowcolor{blue!10}
Heatmap & 39.06 & 39.36 & 17.07 & 32.53 & 31.90 & 11.97 \\

\midrule

\rowcolor{gray!10}
\multicolumn{7}{l}{\textbf{SAM 3}} \\

Upper Bound & 71.18 & 80.06 & 70.96 & 52.92 & 55.78 & 39.50 \\
Gaze-Point & 24.78 & 22.08 & 7.14 & 28.54 & 26.76 & 10.35 \\
\rowcolor{blue!10}
Heatmap  & 30.02 & 27.66 & 13.24 & 35.25 & 35.18 & 18.02 \\
\bottomrule
\end{tabular}
\label{tab:detector_comparison}
\end{table*}
Overall, the results show that the heatmap-based selection strategy is consistently stronger than the single-point gaze baseline across detectors. This trend suggests that using the full gaze heatmap provides a more informative signal than relying only on the peak response. In many cases, the gaze prediction is spatially diffuse rather than concentrated at a single precise point. A heatmap-based score can therefore better capture the region of attention and improve the likelihood of selecting the correct object proposal. This behavior is illustrated in Figure~\ref{fig:heatmap_better}, where the heatmap covers the target object more reliably than the peak gaze point alone, leading to a better bounding-box choice.

\begin{figure}[t]
    \centering
    \includegraphics[width=0.8
    \textwidth]{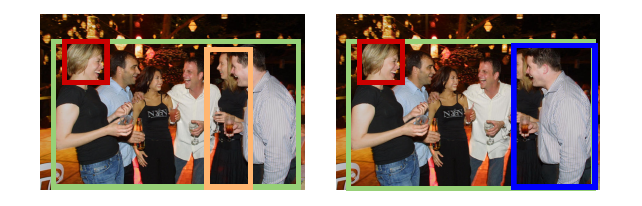}
    \caption{\textbf{Example of annotation ambiguity.} 
    The green box is the ground-truth object, the red box is the head, the orange box is the gaze-point selection, and the blue box is the heatmap-based selection. The prediction is visually reasonable, but the overlap is reduced due to slight ambiguity in the ground-truth annotation.}
    \label{fig:annotation_ambiguity}
\end{figure}

\begin{figure}[t]
    \centering
    \includegraphics[width=0.6
    \textwidth]{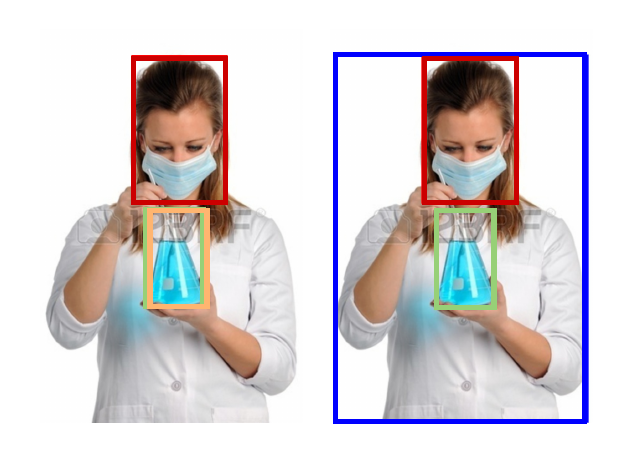}
    \caption{\textbf{Failure case of heatmap-based selection.} 
    The green box is the ground-truth object, the red box is the head, the orange box is the gaze-point selection, and the blue box is the heatmap-based selection. Here, the heatmap is spread across multiple regions, causing the method to choose a larger but less precise bounding box.}
    \label{fig:heatmap_failure}
\end{figure}

At the same time, the heatmap-based method is not universally superior. Because the heatmap may contain multiple activated regions or be spread over a large portion of the image, it can sometimes favor an incorrect proposal, especially a larger box that overlaps several activated areas. In such cases, the richer spatial signal becomes ambiguous rather than helpful. Figure~\ref{fig:heatmap_failure} shows a representative example: the heatmap is distributed across multiple regions, which causes the method to prefer a larger but less precise bounding box over the intended target. This example highlights an important limitation of heatmap scoring when the gaze estimate is fragmented or broadly distributed.

We also observed a small number of cases where the quantitative evaluation may not fully reflect the visual plausibility of the prediction. In particular, some examples appear to contain annotation ambiguity or slight localization imprecision in the ground-truth bounding box. As a result, a prediction that is visually reasonable may receive a lower IoU than expected. Figure~\ref{fig:annotation_ambiguity} presents one such example. We do not view this as a flaw of the dataset, but rather as a natural challenge of large-scale semantic gaze annotation, where object boundaries and gaze targets can occasionally be difficult to define precisely.

\subsection{Prompt Template}
Our framework uses two prompts in sequence. The first prompt is vocabulary-constrained, while the second prompt is retrieval-augmented.

\begin{figure}[h]
    \centering
    \includegraphics[width=0.7\textwidth]{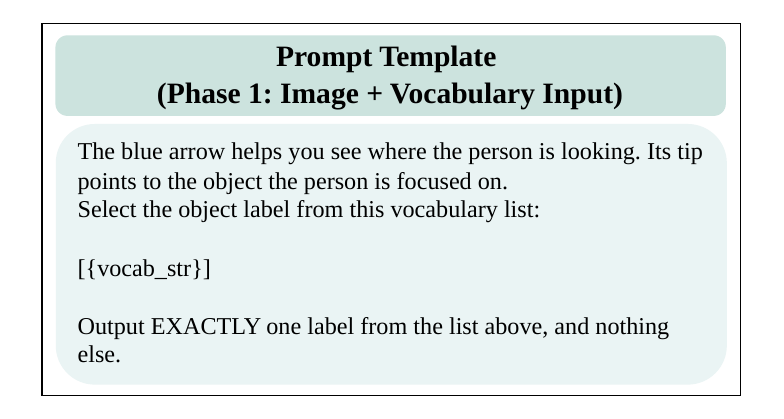}
    \caption{\textbf{Phase 1: Vocabulary-constrained prompt.} The model is given the input image and the benchmark vocabulary, and must select labels only from this predefined label space to ensure fair comparison with prior work.}
\end{figure}

\subsubsection{Phase 1: Vocabulary-Guided Prompt}
In the first stage, the vision-language model is provided with the input image and the full candidate vocabulary. The goal of this prompt is to restrict prediction to the same semantic label space used in prior work. This design ensures a fair comparison with state-of-the-art method by preventing the model from producing unrestricted free-form outputs. The prompt therefore serves as an initial constrained prediction step over the benchmark vocabulary.

\subsubsection{Phase 2: Retrieval-Augmented Prompt}
In the second stage, the model is given a set of retrieved example images with their labels, along with the query image and the candidate labels produced in the first stage. The purpose of this prompt is to use visual analogy and contextual guidance from the retrieved examples to make a more accurate final selection. In this way, the second stage performs a finer decision process over a much smaller and more relevant label space.

\begin{figure}[t]
    \centering
    \includegraphics[width=0.7\textwidth]{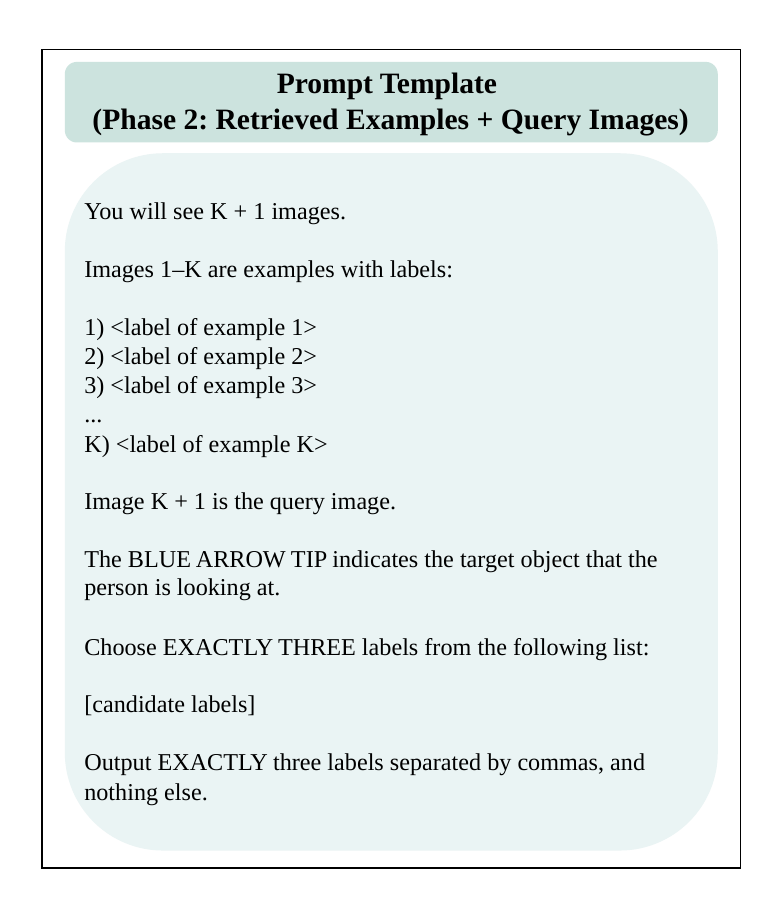}
    \caption{\textbf{Phase 2: Retrieval-augmented prompt.} The model is given the query image, retrieved labeled examples, and the candidate labels from Phase 1 to refine the final gaze-target prediction.}
\end{figure}

\subsection{Additional Qualitative Examples}

In this section, we provide additional qualitative examples to further illustrate the behavior of our framework beyond the main quantitative results. These examples highlight both the strengths of the proposed approach and some of the challenges associated with dataset annotation and constrained evaluation.

\begin{figure}[h]
    \centering
    \includegraphics[width=
    \textwidth]{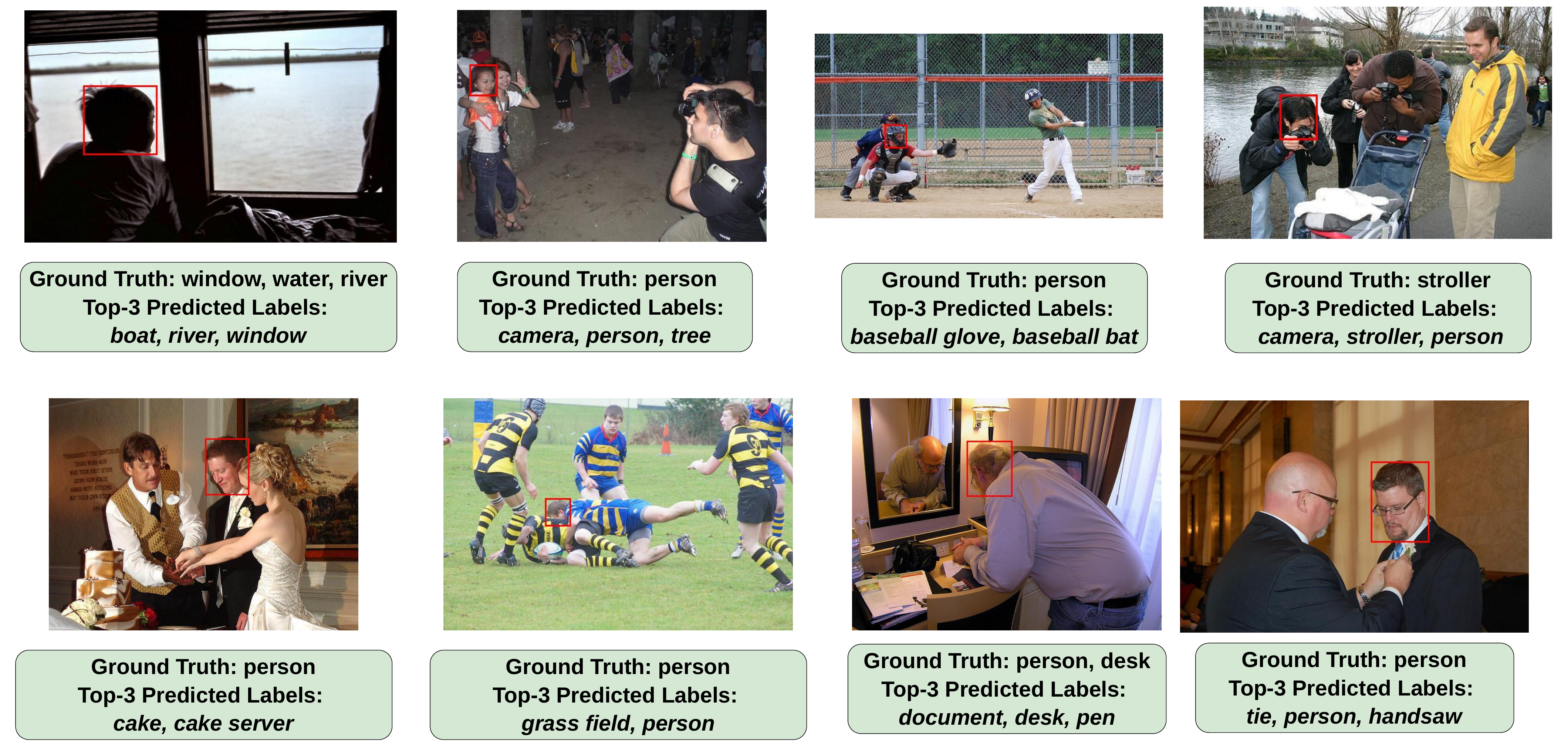}
    \caption{\textbf{Examples of richer semantic predictions.} The agent often predicts labels that are more specific and context-aware than the original dataset annotations, providing a more detailed description of the attended object.}
    \label{fig:semantic_predictions}
\end{figure}

\begin{figure}[t]
    \centering
    \includegraphics[width=0.9
    \textwidth]{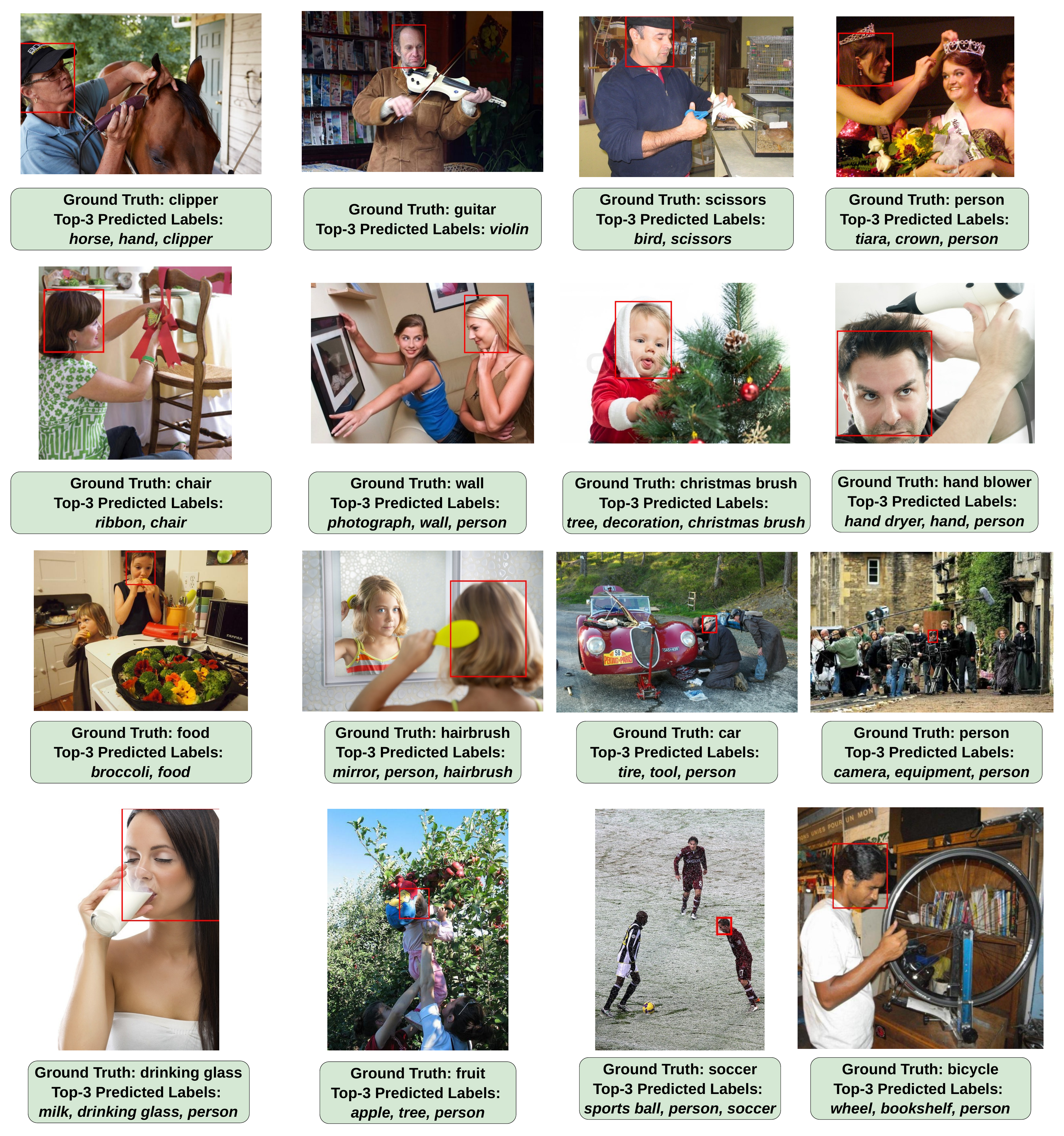}
    \caption{\textbf{Examples of richer semantic predictions.} The agent often predicts labels that are more specific and context-aware than the original dataset annotations, providing a more detailed description of the attended object.}
    \label{fig:semantic_predictions1}
\end{figure}

Figures~\ref{fig:semantic_predictions} and~\ref{fig:semantic_predictions1} show cases where the predicted labels provide a more detailed semantic description of the attended object than the original dataset annotation. Instead of producing only generic object names, the agent often outputs labels that better capture the visual context and the semantic role of the object in the scene. This suggests that the framework is able to exploit the semantic knowledge of vision-language models to generate richer gaze-target interpretations.

\begin{figure}[h]
    \centering
    \includegraphics[width=0.9
    \textwidth]{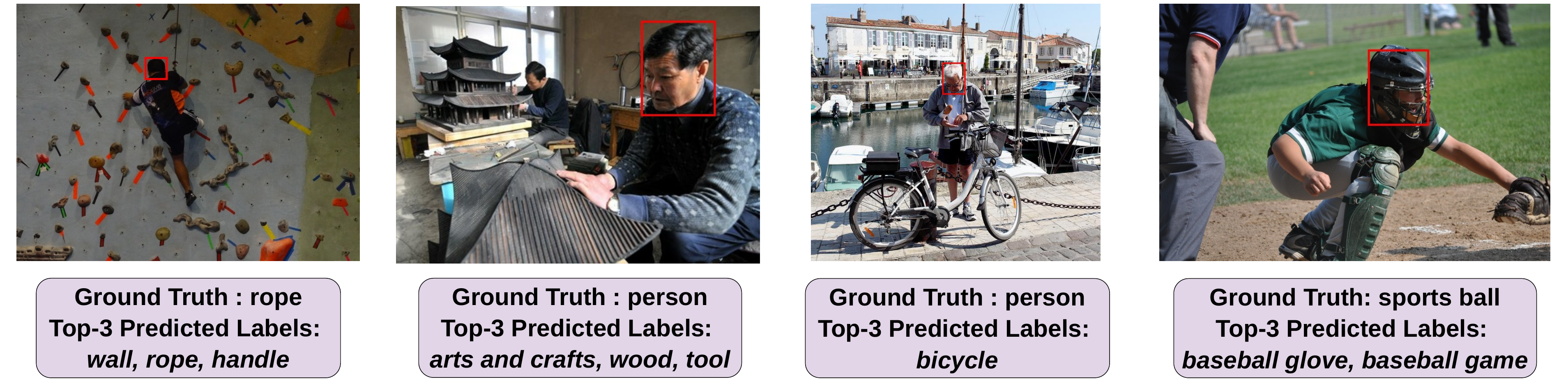}
    \caption{\textbf{Examples of annotation ambiguity.} In some cases, the dataset annotation appears simplified or slightly inconsistent with the visual scene, while the agent prediction remains visually plausible and semantically informative.}
    \label{fig:gt_errors}
\end{figure}

\begin{figure}[t]
    \centering
    \includegraphics[width=\textwidth]{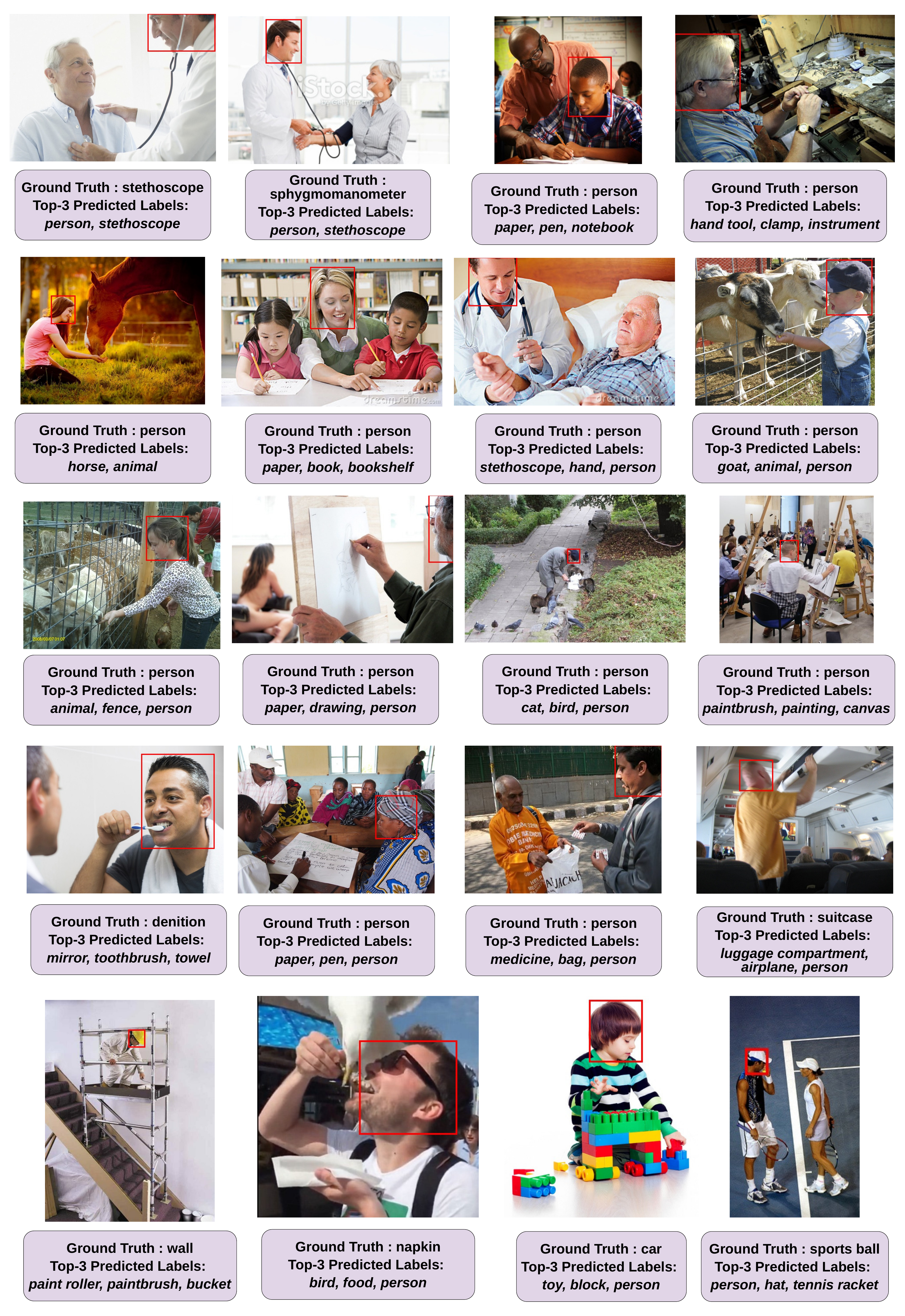}
    \caption{\textbf{Examples of annotation ambiguity.} In some cases, the dataset annotation appears simplified or slightly inconsistent with the visual scene, while the agent prediction remains visually plausible and semantically informative.}
    \label{fig:gt_errors1}
\end{figure}

Figures~\ref{fig:gt_errors} and~\ref{fig:gt_errors1} present examples where the dataset annotation appears ambiguous, simplified, or slightly misaligned with the visual evidence. In these cases, the agent predicts labels that appear more consistent with the actual gaze target. These examples highlight the inherent difficulty of large-scale semantic gaze annotation and suggest that model predictions can sometimes expose annotation inconsistencies or borderline cases in the benchmark.

\begin{figure}[t]
    \centering
    \includegraphics[width=\textwidth]{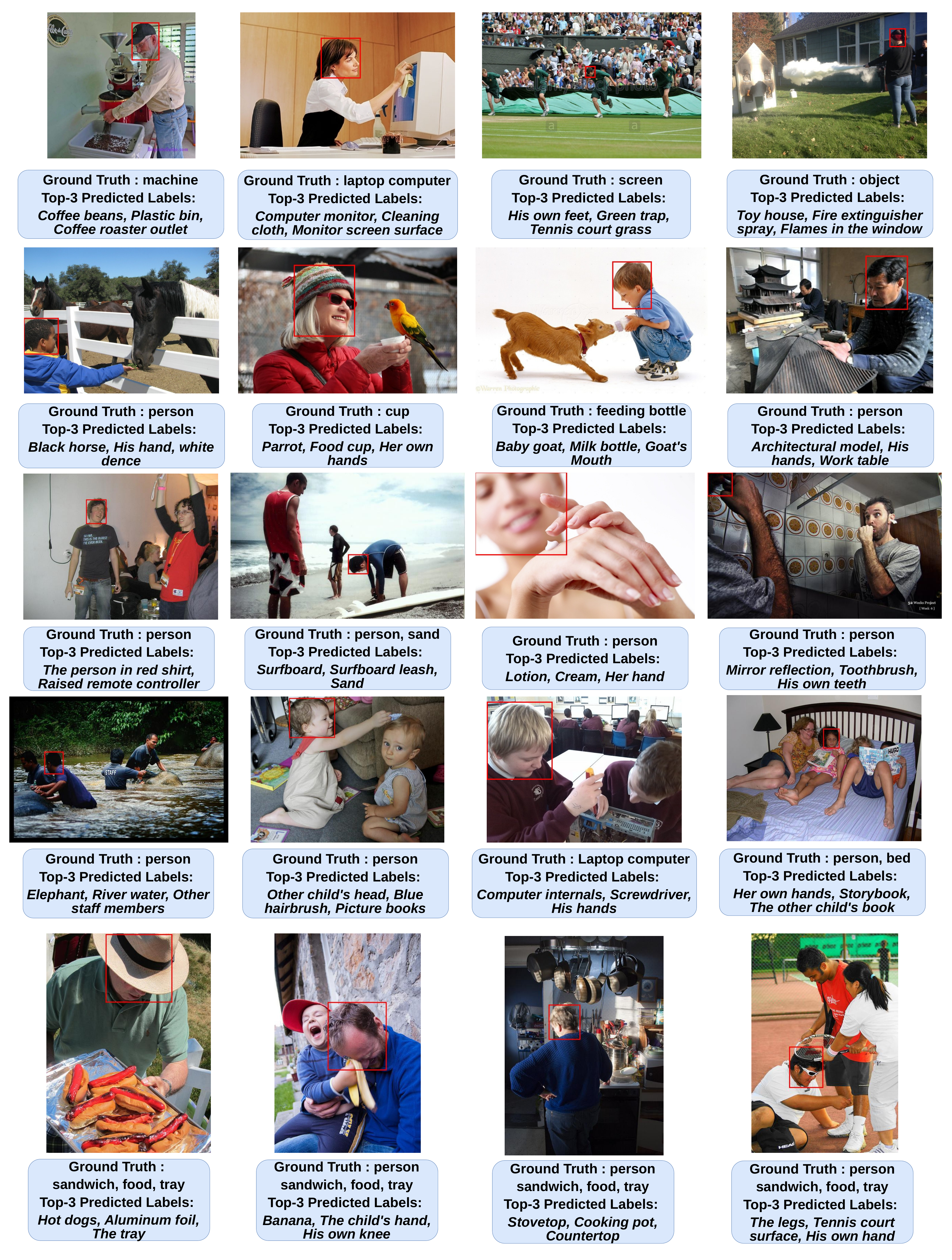}
    \caption{\textbf{Examples of open-vocabulary predictions.} Without restricting the output to a predefined vocabulary, the agent can generate more flexible and semantically rich descriptions of the attended object.}
    \label{fig:open_predictions}
\end{figure}

Finally, Figure~\ref{fig:open_predictions} illustrates the behavior of the agent when prediction is performed without restricting the output to a predefined vocabulary. In this open-vocabulary setting, the model often generates more precise and context-aware object descriptions than those allowed by the fixed benchmark label space. These examples show that the agent can leverage open-vocabulary reasoning to produce flexible and semantically informative gaze-target predictions beyond the categories used in quantitative evaluation.

